\documentclass[10pt,twocolumn,letterpaper]{article}

\usepackage[pagenumbers]{iccv} 

\definecolor{iccvblue}{rgb}{0.21,0.49,0.74}
\usepackage[pagebackref,breaklinks,colorlinks,allcolors=iccvblue]{hyperref}

\def\paperID{9312} 
\def\confName{ICCV}
\def\confYear{2025}

\title{Inversion-Free Video Style Transfer with Trajectory Reset Attention Control and Content-Style Bridging}

\author{Jiang Lin\\
Nanjing University\\
 {\tt\small 602024710007@smail.nju.edu.cn}
\and
 Zili Yi\\
Nanjing University\\
}

\begin{document}
\maketitle
\begin{figure*}[t!] 
\centering 
\includegraphics[width=\linewidth]{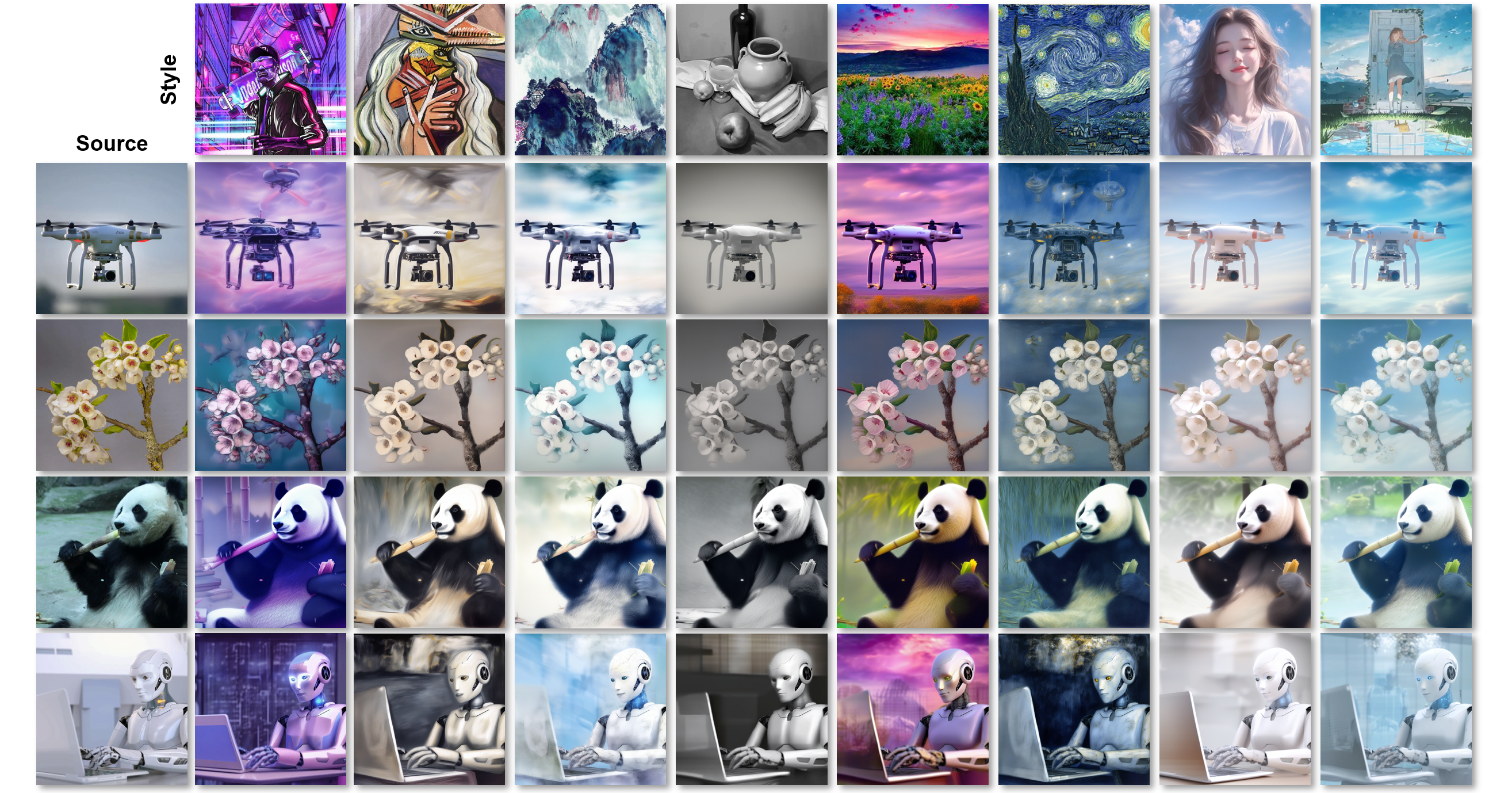} 
\caption{Visualizations on Image style transfer.}
\label{fig: qualitative}
\end{figure*}

\begin{abstract}

Video style transfer aims to alter the style of a video while preserving its content. Previous methods often struggle with content leakage and style misalignment, particularly when using image-driven approaches that aim to transfer precise styles. In this work, we introduce \textbf{ Trajectory Reset Attention Control (TRAC)}, a novel method that allows for high-quality style transfer while preserving content integrity. TRAC operates by resetting the denoising trajectory and enforcing attention control, thus enhancing content consistency while significantly reducing the computational costs against inversion-based methods. Additionally, a concept termed \textbf{Style Medium} is introduced to bridge the gap between content and style, enabling a more precise and harmonious transfer of stylistic elements. Building upon these concepts, we present a tuning-free framework that offers a stable, flexible, and efficient solution for both image and video style transfer. Experimental results demonstrate that our proposed framework accommodates a wide range of stylized outputs, from precise content preservation to the production of visually striking results with vibrant and expressive styles.

\end{abstract}

\section{Introduction}
\label{sec:intro}

Style transfer tasks have gained significant attention in artistic design, entertainment, and social media, allowing users to transform the appearance of images and videos while preserving their content. While image style transfer focuses on achieving static style transformations, video style transfer introduces an additional challenge: ensuring temporal consistency across frames. Traditional methods for style transfer, often data-driven and domain-specific \cite{Zhu_2017_ICCV}, are increasingly being supplanted by diffusion models. These methods offer greater flexibility and generalization, allowing content to be adapted into arbitrary styles based on image or text references, thereby providing broader applicability.

Diffusion-based style transfer methods can be broadly divided into text-driven \cite{he_freestyle_2024,jiang_artist_2024} and image-driven \cite{ye_ip-adapter_2023,wang_instantstyle_2024,gao_styleshot_2024} approaches. Text-driven methods use detailed prompts to guide style transfer but often struggle with language ambiguity and difficulties in capturing niche or complex styles. In contrast, image-driven methods offer more precise style transfer by utilizing reference images but introduce their own challenges, primarily content leakage. Content leakage occurs when non-style elements from the reference image inadvertently infiltrate the generated content, leading to distortions in the target image. Although some methods attempt to mitigate this issue using style encoders \cite{gao_styleshot_2024} or feature injection control mechanisms \cite{wang_instantstyle_2024},  they still seek fully disentanglement of style and content. With the content being fully disentangled, the cross-attention mechanism fails to associate the style information with the fragmented information, resulting in style mismatches.


Besides, to provide explicit control on the content, previous methods often employ DDIM inversion \cite{song2022denoisingdiffusionimplicitmodels}, but they are prone to reconstruction errors while incurring high computational costs. Additionally, inversion-based video methods rely on a suitable generic text-to-video diffusion model, which current approaches fail to perform satisfactorily.
To tackle these challenges, we introduce Trajectory Reset Attention Control (TRAC), a novel inversion-free method that preserves content integrity while enabling high-quality style transfer. Unlike traditional methods that fully disentangle style and content, TRAC enables direct style infusion while preserving the original content structure, effectively preventing content leakage.

Specifically, TRAC introduces an auxiliary denoising path and resets the intermediate latent at each timestep, ensuring the diffusion trajectory aligns with the ``ideal'' path and effectively preserving content information. This content information is subsequently extracted from self-attention matrices which are then used to retain content and layout in the main diffusion process. Unlike inversion-based methods \cite{song2022denoisingdiffusionimplicitmodels,gu_photoswap_2023} that operate on inverted noises, TRAC directly utilizes the noised latents generated by the forward diffusion process \cite{meng2022sdeditguidedimagesynthesis}, which significantly reduce additional computation cost.



Additionally, to overcome the issue of style misalignment, we introduce a novel concept termed \textit{Style Medium}. This intermediary representation combines the content from the target video with the style extracted from the reference image, ensuring that the style elements align with the corresponding content elements in the video frames. Specifically, a Multimodal Large Language Model (MLLM) is used to extract textual descriptions from the video frames,  which is further used as prompts to guide the creation of the Style Medium image with target style features infused.
The Style Medium acts as an intermediary between style and content, ensuring that the style from the reference image is accurately applied to the corresponding semantic elements in the target video frames. While TRAC manages the preservation of the content structure, the Style Medium ensures that style information is seamlessly and correctly integrated into the target video, preventing style misalignment. Furthermore, it significantly reduces content leakage, alleviating the burden on TRAC and resulting in better content preservation through their joint efforts.

Building upon these components, we develop a video style transfer framework that ensures high content consistency and precise style matching. To effectively balance structural details with stylization authenticity, we also integrate ControlNet \cite{zhang2023addingconditionalcontroltexttoimage} into the framework, which utilizes canny edge information to restrict structural details, thus allowing for a nuanced trade-off between maintaining structural integrity and achieving authentic stylization. Our video style transfer framework consistently delivers precisely stylized results that preserve content integrity, as demonstrated through extensive experiments. Our main contributions can be summarized as:
\begin{itemize}
\item We propose TRAC, an efficient attention control method operates without DDIM inversion. By proposing a trajectory reset mechanism, it prevents cumulative deviation, and enables precise style infusion while preserving content integrity.
\item We introduce the novel concept of Style Medium, an intermediary that ensures seamless and correct application of style elements to the target while preserving the original content structure.
\item 
We develop a comprehensive Video Style Transfer Framework that delivers high-quality, visually coherent stylized videos with improved flexibility and robustness, as demonstrated through extensive experiments.

\end{itemize}


\section{Related Works}
\label{sec:related}

Diffusion-based methods \cite{ahn_dreamstyler_2023, xu_freetuner_2024, he_freestyle_2024} enable more flexible style transfer guided by images or text.


\textbf{Text-Driven Methods} Text-driven methods \cite{he_freestyle_2024, jiang_artist_2024} enable style transfer using textual descriptions, eliminating the need for style images. Large pre-trained models \cite{rombach2022highresolutionimagesynthesislatent} have acquired sufficient capability during their training to connect visual styles with text prompts, allowing them to generate images in various styles based on text inputs. However, these methods rely on the stylistic information within pre-trained models, limiting their ability to generalize to less common styles. Text prompts often lack the precision needed for accurate style representation, making prompt engineering cumbersome compared to the more intuitive use of image-based style references.

\textbf{Image-Driven Methods} Image-driven methods draw style information from the given reference images. Fine-tuning approaches \cite{sohn_styledrop_2023, karim_save_2023, wang_stylediffusion_2023} adjust the model based on a set of reference images, but they face challenges when fine-tuning with small datasets. To utilize image information without tuning, IP-Adapter \cite{ye_ip-adapter_2023} introduces a decoupled cross-attention mechanism that integrates image features into the denoising process. However, this approach introduces content information along with style, potentially distorting the structure of the stylized image. Recent efforts \cite{wang_stylediffusion_2023, jiang_artist_2024, wang_instantstyle_2024} have aimed to disentangle style from content by extracting style representations from images, resulting in better content preservation. Nevertheless, this approach may result in the inaccurate fusion of style with corresponding elements, leading to inconsistencies.


\textbf{Video Style Transfer} 
Previous methods \cite{chen2017coherent, gao2020fast, xia2021real, huang2017real} in video style transfer incorporate optical flow constraints to enhance inter-frame correlation, but their high computational cost limits scalability for high-resolution or long-duration videos. Attention-based models \cite{deng2021arbitrary, liu2021adaattn, wu2020preserving} improve temporal consistency but require fine-tuning. Both image and video style transfer also struggle with balancing content consistency and style strength \cite{jeong2024visual, qi2024deadiff}. 


\textbf{Controllable Diffusion Generation}
Maintaining content consistency in style transfer is challenging, as it requires preserving original content while integrating style information. SDEdit \cite{meng2022sdeditguidedimagesynthesis} retains low-frequency content by adding noise to the original image, where increased noise enhances stylization but also weakens content preservation. ControlNet \cite{zhang2023addingconditionalcontroltexttoimage} introduces an auxiliary network for conditioning in the diffusion process but prioritizes layout over content, leading to varying preservation quality. Other works \cite{wang_stylediffusion_2023, karim_save_2023} apply DDIM inversion \cite{song2022denoisingdiffusionimplicitmodels} to inject style during denoising, with auxiliary paths supplying content control through attention or feature injection \cite{jiang_artist_2024, gu_photoswap_2023}.



\section{Method}
\label{sec: method}

\begin{figure}[htb] 
\centering 
\includegraphics[width=\linewidth]{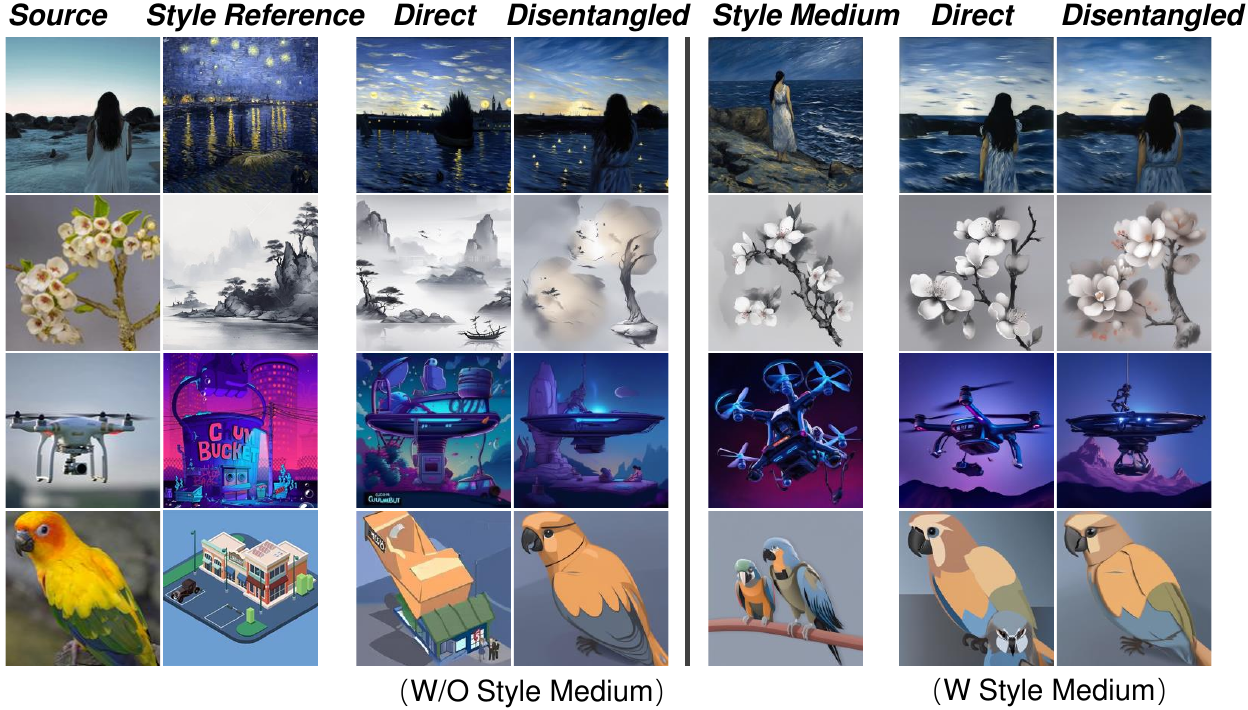} 
\caption{ Comparison on the effect of using Style Medium.}
\label{fig: style medium}
\end{figure}

\begin{figure*}[htb] 
\centering 
\includegraphics[width=\textwidth]{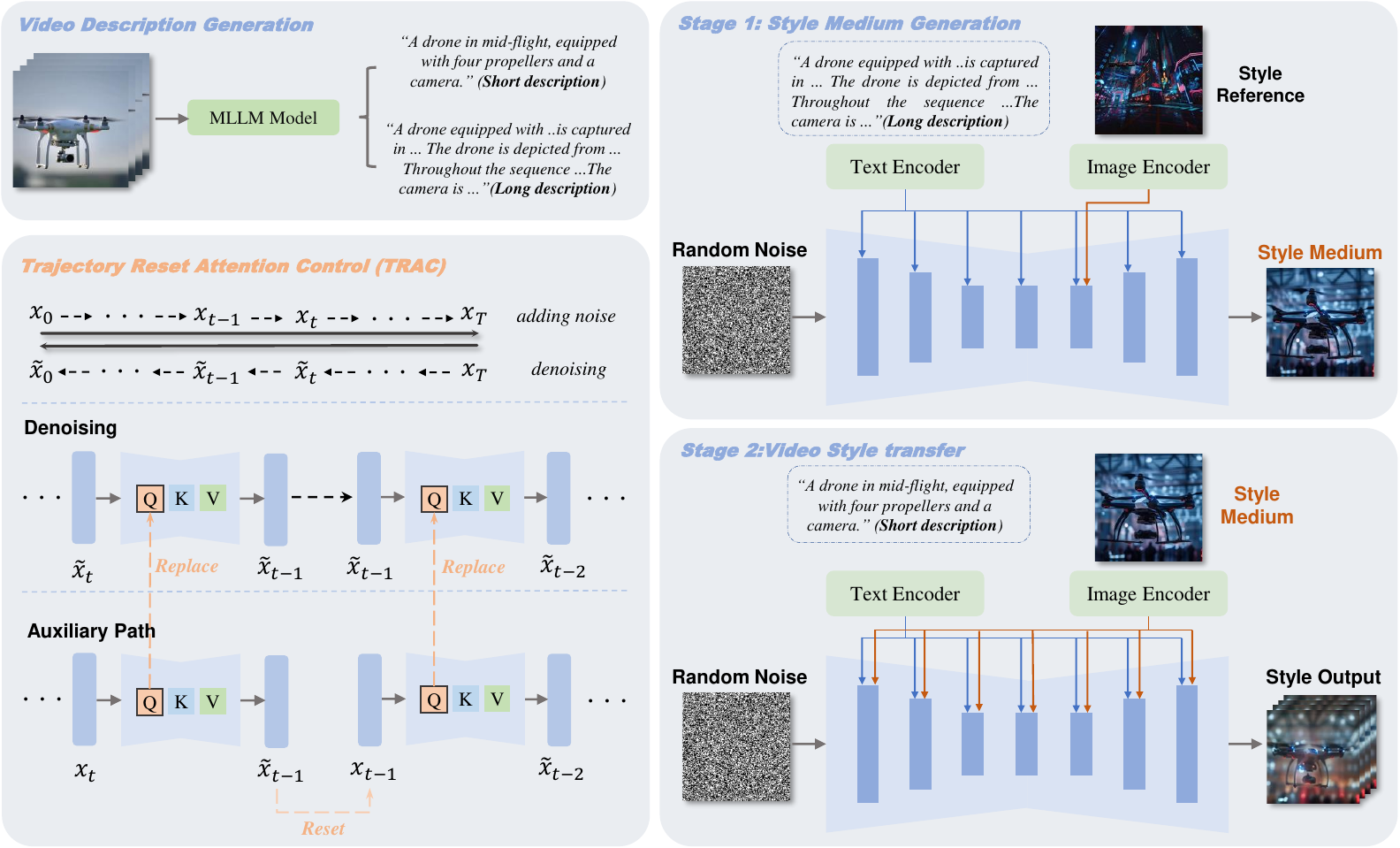} 
\caption{
The video style transfer framework operates as follows: The source video is first processed by an MLLM model to generate two descriptions. The long description guides the content, while the style-specific features from the style reference inform the generation of the style medium. This style medium then serves as the new style reference for the video style transfer process. Meanwhile, the short description aids in generating the final stylized video.}
\label{fig: framework}
\end{figure*}

\subsection{Style Medium}
In this section, we propose a method to generate a \textit{style medium}, \( I_{sm} \), that bridges the style transfer between a style reference image, \( I_s \), and a content video, while effectively preventing content leakage. This method addresses the challenge of achieving style fidelity while preserving the content of the video.

Selecting an appropriate style reference is crucial in the style transfer task. Consider two style images, \( I_s \) and \( I_d \), which share the same visual style but differ in their content similarity to the source video. When these images are introduced into the diffusion process through cross-attention, as illustrated in \cref{fig: style medium}, the style image \( I_s \) with greater content similarity to the source video results in better alignment. Conversely, \( I_d \), which differs in content, leads to content leakage and undesirable artifacts. However, finding a style image that closely matches the content of the source video is often difficult due to data limitations, and identifying a suitable reference can be a time-consuming process.

To address this challenge, we introduce the concept of a \textbf{style medium}, \( I_{sm} \), an image that retains the content from the source video while adopting the style of the reference image \( I_s \). This style medium is generated through disentangled style encoding \cite{wang_instantstyle_2024}, which ensures that only style-specific features are injected into the model. To provide precise content guidance during generation, we further employ an MLLM model \cite{yao2024minicpmvgpt4vlevelmllm} to extract key elements from the source video and construct a descriptive text prompt. The prompt is later used to control the content of the style medium and ensure it includes all crucial elements. 
This approach allows image style guidance and text content guidance to operate in parallel, gradually merging as the diffusion process progresses. Examples of the style medium can be seen in \cref{fig: style medium}.

The style medium effectively captures the target style while maintaining content relevance and does not need to have layout consistency with the source video, as each frame naturally varies with motion. As long as the style medium possesses sufficient content similarity, the cross-attention mechanism will align it with the corresponding elements, ensuring it serves as an appropriate candidate for style reference.


\begin{figure}[htb] 
\centering 
\includegraphics[width=\linewidth]{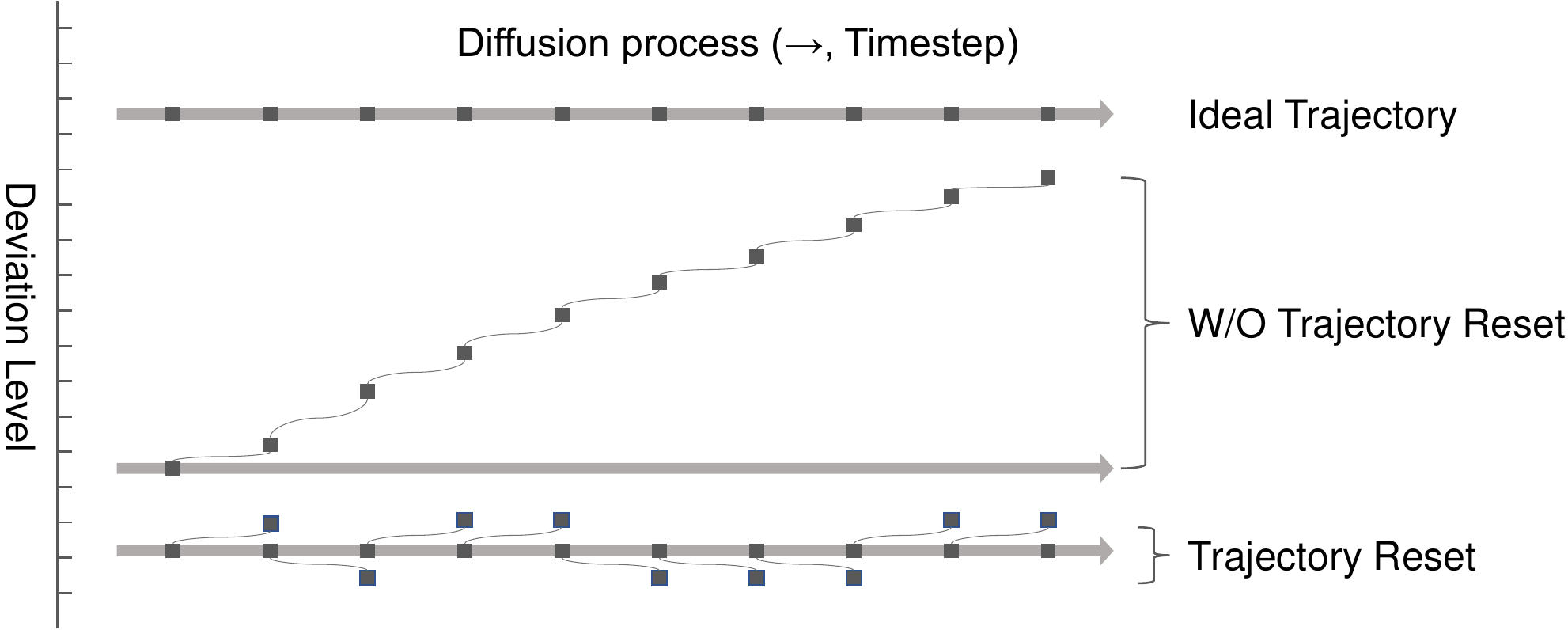} 
\caption{Illustration of the prediction deviation as the diffusion process advances. }
\label{fig: deviate}
\end{figure}

\subsection{Trajectory Reset Attention Control (TRAC)}
While style mediums help mitigate content leakage, they do not offer an explicit mechanism to ensure consistency in content and layout. In this section, we introduce \textbf{Trajectory Reset Attention Control (TRAC)}—a novel method that preserves the original content and layout of a video while allowing high-quality style transfer.

In previous methods \cite{hertz_prompt--prompt_2022,gu_photoswap_2023}, different components within the attention calculation have been shown to control various aspects of the generation process. In particular, the query matrix (Q) in the self-attention layer plays a crucial role in managing both layout and content. To preserve the original content in the generated video, we use a straightforward approach: replacing the query matrix with its corresponding counterpart that retains information from the original video.

As illustrated in \cref{fig: framework}, TRAC constructs an auxiliary path that supplies essential content information to the main path at each timestep of the denoising process. Specifically, in each self-attention layer of the auxiliary path, the query matrix—responsible for encapsulating content details—is extracted at each timestep and injected into the self-attention mechanism of the main path. Formally, if \( Q_{\text{aux}}(t) \) denotes the query matrix extracted from the auxiliary path at timestep \( t \) and \( Q_{\text{main}}(t) \) is the query matrix in the main path (layer is ignored for simplicity), we enforce:
\[
Q_{\text{main}}(t) \leftarrow Q_{\text{aux}}(t)
\]
This replacement ensures that the content information is consistently preserved throughout the denoising process.

To construct a path that carries the information of the original video, previous methods often rely on techniques such as DDIM Inversion \cite{song2022denoisingdiffusionimplicitmodels} to recover the original latent. However, these methods are computationally expensive and suffer from reconstruction errors. Instead of using an inverted latent, TRAC operates on the noised latent generated using SDEdit \cite{meng2022sdeditguidedimagesynthesis}. 
Under this setting, the denoising process refines a noised latent \( x_t \) toward its original content \( x_0 \). For the extracted query matrix to contain the correct content information, it is crucial for the denoising trajectory to traverse back to the original video latent; otherwise, the content information will be distorted. In the SDEdit setting, the noised latent does not naturally return to the original latent \(x_0\) as it does in DDIM inversion. As the denoising process progresses, the latent \( x_t \) often diverges from the ideal trajectory, as shown in the middle of \cref{fig: deviate}. 

To address this issue, TRAC introduces a trajectory reset mechanism. Rather than relying on inversion techniques, TRAC resets the intermediate latent, denoted by \(\tilde{x}_t\), at each timestep to align it with the ideal diffusion trajectory—the path that brings the noised latent \(x_t\) back to the original latent \(x_0\). The reset operation is defined as follows:
\[
\tilde{x}_t \leftarrow \text{forward}(x_0, \epsilon, t)
\]
Here, \( \epsilon \) represents the randomly generated noise used in the forward process.
As depicted in \cref{fig: deviate}, the denoised latent is replaced at every timestep with the latent points from the ideal trajectory (generated by the forward noise process).

The core assumption behind this method is that the predicted latent \( \tilde{x}_{t-1} \), computed from \( x_t \), approximates the ideal latent point \( x_{t-1} \). While the divergence introduced by one prediction is small, if the process continues on a diverged latent, the accumulated error can drive the result away from \( x_0 \). In other words, with a diverged input \( \tilde{x}_t \) instead of \( x_t \), the predicted noise \( \tilde{\epsilon} \) becomes more divergent, which in turn produces a more deviated \( \tilde{x}_{t-1} \). In high dimensional spaces, every divergence almost always drives the predicted results further away from the ideal trajectory.

We conjecture that latent points on the ideal trajectory have the highest probability of being restored to the original latent \( x_0 \) during denoising and thus provide the most reliable content information. By constantly resetting the trajectory, we ensure that the predicted latent points at every timestep diverge only slightly, thereby closely approximating the ideal trajectory. This reset operation prevents the latent from straying too far and guarantees that the content information is preserved throughout the denoising process.

By injecting the content information from the auxiliary path into the main path, TRAC successfully maintains content integrity. Moreover, TRAC offers a significant computational advantage, as the reset operation incurs negligible expense compared to inversion-based methods.

\subsection{Video Style Transfer Framework}
Our proposed video style transfer framework, illustrated in \cref{fig: framework}, consists of three stages: description generation, style medium generation, and video style transfer.
In the first stage, an MLLM model \cite{yao2024minicpmvgpt4vlevelmllm} generates two descriptions of the original video—one detailed and one concise—both excluding style-related elements. Next, a style image \( I_s \) is processed into a style medium \( I_m \). In the transfer stage, we employ Stable Diffusion \cite{rombach2022highresolutionimagesynthesislatent} as our base model, integrating AnimateDiff's Motion Adapter \cite{guo_animatediff_2024} within a 3D-UNet \cite{ho2022videodiffusionmodels} to enhance motion consistency. The style medium \( I_m \) is injected via cross-attention using the IP-Adapter \cite{ye_ip-adapter_2023}, alongside an auxiliary path with TRAC to enforce content and layout control.
For finer structural details, we also incorporate ControlNet \cite{zhang2023addingconditionalcontroltexttoimage}, using Canny edge extracted from the frames of \( V_c \). This step, omitted in \cref{fig: framework} for simplicity, further refines video structure during the denoising process.

The motion adapter from AnimateDiff \cite{guo_animatediff_2024} enhances frame smoothness but is insufficient for resolving content variability across frames. TRAC addresses this limitation by regulating content and layout consistency in each frame, effectively mitigating issues such as flickering and shape distortion.  Furthermore, we integrate TRAC within motion modules to reinforce motion alignment with the original video, further improving content coherence throughout the diffusion process. By ensuring spatial alignment with the original video, which is presumably temporal consistent, TRAC also enhances temporal stability.


\section{Experiments}
\label{sec:exp}


\subsection{Implementation Details}
Our style transfer framework is built on Stable Diffusion 1.5 and SDXL with AnimateDiff \cite{guo_animatediff_2024}. We use IP-Adapter \cite{ye_ip-adapter_2023} for image feature injection in both style medium generation and video style transfer. Text prompts are generated by MiniCPM \cite{yao2024minicpmvgpt4vlevelmllm}.

\subsection{Image Style Transfer}
For quantitative comparison, we adopt metrics such as LPIPS, FID and ArtFID \cite{wright_artfid_2022}.
ArtFID is computed as \(ArtFID = (1 + LPIPS)\times(1 + FID)\). LPIPS measures content fidelity, while FID assesses the style fidelity.
Follow previous methods \cite{deng_stytr2_2022}, we employ content images from MSCOCO \cite{lin_microsoft_2015} dataset and style images from WikiArt \cite{phillips2011wiki} dataset. A total of  20 content and 40 style images are drawn from each dataset, yielding 800 stylized images.




\begin{table}
\centering
\small
\resizebox{\linewidth}{!}{
\begin{tabular}{c c c c c c}
\hline
Metric & Ours  &  InstantStyle & DiffuseIT  & InST & StyleID \\ 
\hline
\hline
 $\downarrow$ ArtFID & 36.630 &  37.524 & 40.721 & 40.633 & 28.801 \\ 
 $\downarrow$ FID & 24.415 & 21.817 & 23.065 & 21.571 &  18.131\\ 
 $\downarrow$ LPIPS & 0.4413  & 0.6446  & 0.6921 & 0.8002  & 0.5055  \\ 
 \hline 
 \\
\end{tabular}}
\caption{Quantitative comparison against existing image style transfer methods.}
\label{tab: image_metric}
\end{table}

\textbf{Quantitative Comparison}
We conduct quantitative evaluations with previous methods, including StyleID \cite{Chung_2024_CVPR_style_ID}, InstantStyle\cite{wang_instantstyle_2024}, DiffuseIT\cite{kwon_diffusion-based_2023} and InST\cite{zhang_inversion-based_2023}. As shown in \cref{tab: image_metric}, our method outperforms them in LPIPS, which suggests higher content alignment. The FID results are suboptimal, likely because previous methods often experience content leakage, introducing elements from the style reference into the final output. This increases correlation between the final output and the style reference, leading to better FID scores. Additionally, this dataset, predominantly used in previous evaluations, consists mainly of artistic paintings, potentially introducing bias in the assessments. Therefore, we provide these results for reference only.

\begin{figure}
\centering 
\includegraphics[width=\linewidth]{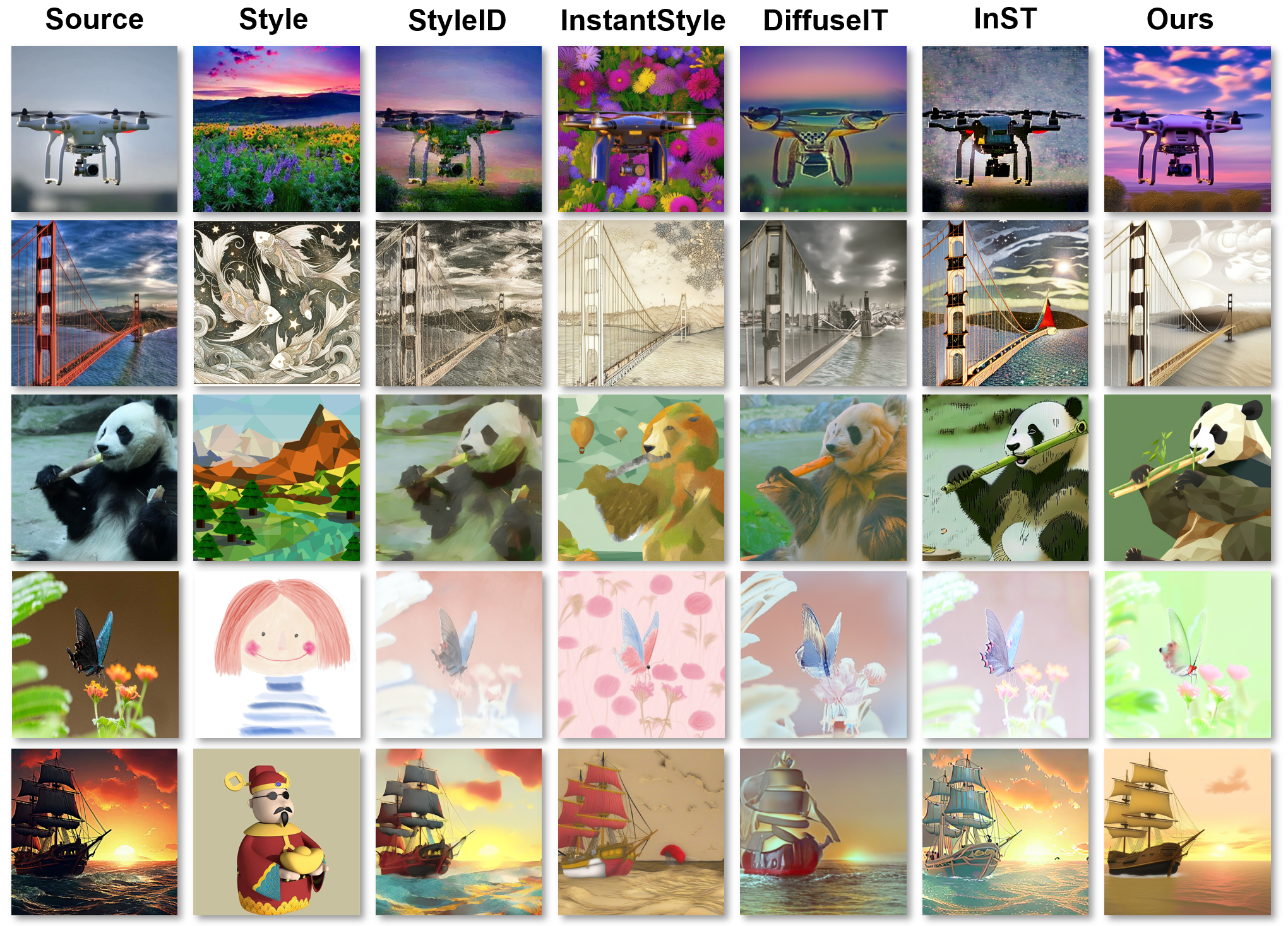} 

\caption{Comparison of image style transfer results.}
\label{fig: image comparison}
\end{figure}

\textbf{Qualitative Comparison}
\cref{fig: image comparison} presents a visual comparison between our method and other state-of-the-art approaches. Overall, our method achieves the optimal visual balance between enhancing stylistic effects and preserving the original content, while effectively preventing content leakage from the style image. Several key observations can be made from this figure:
First, our methods that do not incorporate inversion or introduce a style medium exhibit significant limitations in content preservation, especially in background details.
Second, although inversion-based methods such as StyleID, InST, and InstantStyle demonstrate some ability to preserve content, they fail to fully decouple style and content information, which results in visible content leakage in some synthesized images.

\begin{table}
\centering
\small
\begin{tabular}{l c c c c }
\hline
Metric & CAP-VST  &  UniST & IP+ANI  & Ours \\ 
\hline
\hline
 $\downarrow$ LPIPS &  0.179&0.286 & 0.482 &    0.31 \\ 
 $\uparrow$ CLIP &  0.920 & 0.842 & 0.626 &  0.85\\ 
 $\uparrow$ MLLM(C) & 8.23 &7.95  & 7.01 & 7.47    \\ 
 $\uparrow$ MLLM(S) &7.82 & 7.75 & 7.90 & 8.21   \\ 
 \hline 
 \\
\end{tabular}
\caption{Quantitative comparisons with existing video style transfer methods.}
\label{tab: metric}
\end{table}

\begin{figure*}
\centering 
\includegraphics[width=\textwidth]{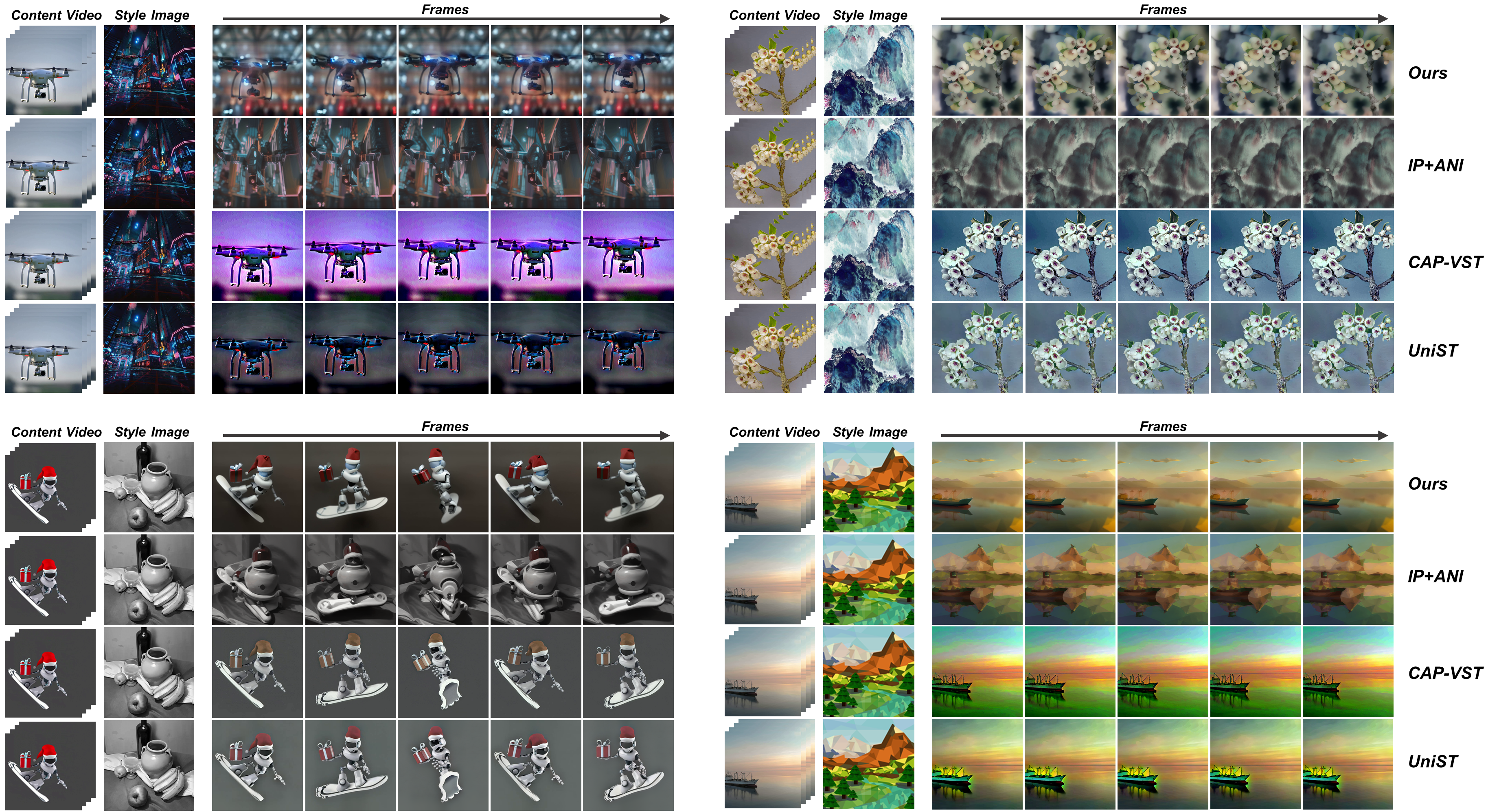} 

\caption{Comparison of video style transfer results.}
\label{fig: comparison}
\end{figure*}

\subsection{Video style transfer}
\textbf{Quantitative Comparison} 
 We compare our approach with CAP-VST \cite{wen_cap-vstnet_2023}, UniST \cite{gu2023birdsstoneunifiedframework}, and a combination of AnimateDiff \cite{guo_animatediff_2024} and IP-Adapter \cite{ye_ip-adapter_2023}. Evaluation metrics include LPIPS, CLIP cosine similarity, and  we additionally exploit an MLLM\cite{yao2024minicpmvgpt4vlevelmllm} for assessing style fidelity and content alignment. While existing metrics are not fully sufficient for evaluating video style transfer quality, we include them for completeness.






The results in \cref{tab: metric} show that while CAP-VST and UniST achieve higher LPIPS and CLIP scores, they receive lower style scores from the MLLM model. As seen in \cref{fig: comparison}, their strong content alignment stems from insufficient stylization. In contrast, our method, despite lower content alignment scores, achieves better style fidelity per the MLLM model, indicating a trade-off where slight structural loss is necessary for authentic style transfer.

\textbf{Qualitative Comparison}
A visual comparison with baseline methods is shown in \cref{fig: comparison}. CAP-VSTNet \cite{wen_cap-vstnet_2023} preserves content well but applies minimal style transfer, mainly altering color and lighting. UniST \cite{gu2023birdsstoneunifiedframework} achieves stronger stylization but introduces instability and flickering. Both methods primarily modify color and contrast, resulting in an artificial filter-like effect while suffering from frequent anomalies such as overexposure and excessive contrast, degrading visual quality. As a training-free approach, our method achieves superior stylization quality compared to previous training-dependent methods.

For the previous close-related competitor, while style similarity is higher, content preservation is poor, leading to severe distortion and a lack of authenticity. In contrast, our method balances content alignment and style fidelity, maintaining the original layout while incorporating stylistic details. The content remains consistent across frames without significant distortion or transformations. Additional examples in the supplementary material further illustrate these differences.

\begin{table}
\centering
\begin{tabular}{c c c c c }
\hline
Metric & CAP-VST &  UniST  & Ours   \\ 
\hline
\hline
Content alignment & 7.38 & 7.25 & 6.65    \\ 
Style transfer quality & 6.33 & 6.12 & 6.94    \\ 
Overall preference & 7.18 & 6.87 & 7.89    \\ 
 \hline 
 \\
\end{tabular}
\caption{User study against existing video style transfer methods.}
\label{tab: user}
\end{table}

\subsection{User Study}
To enhance the representativeness of our experiments, we conducted a user study to evaluate video stylization quality. Using 20 content videos and 20 style images, we generated 400 stylized videos. Forty participants were invited to rate 10 randomly assigned video-style pairs, each containing results from three methods, including ours. Ratings were based on content alignment, style transfer quality, and overall preference. As shown in \cref{tab: user}, our method was preferred in style transfer quality and overall preference.

\begin{table}[bht]
\centering
\begin{tabular}{l r r}
\hline
        & Ours  & Inversion \\ 
\hline
\hline
Time per Iteration (Seconds) & 69.55 & 90.15    \\ 
 \hline 
 \\
\end{tabular}
\caption{Comparison of Average Inference Time per test case (20-frame Video at 512$\times$512 Resolution)  on NVIDIA RTX A6000 GPU of Our Method and Inversion-based method.}
\label{tab: efficiency}
\end{table}

\subsection{Efficiency Analysis}

Our attention control method introduces additional forward noise operations equal to the number of denoising steps, followed by the corresponding denoising process. In contrast, the DDIM inversion auxiliary path involves a series of inversion operations, also followed by denoising.
The majority of computational time in inversion-based methods is spent on UNet prediction \cite{ronneberger2015unetconvolutionalnetworksbiomedical}, requiring an extra prediction at each timestep. Although existing models struggle with proper video latent inversion, we simulate inference times in \cref{tab: efficiency}, demonstrating that our attention-based method is faster than inversion-based approaches.


\subsection{Ablation}
\label{sec:ablation}



\begin{figure*}[htb] 
\centering 
\includegraphics[width=\textwidth]{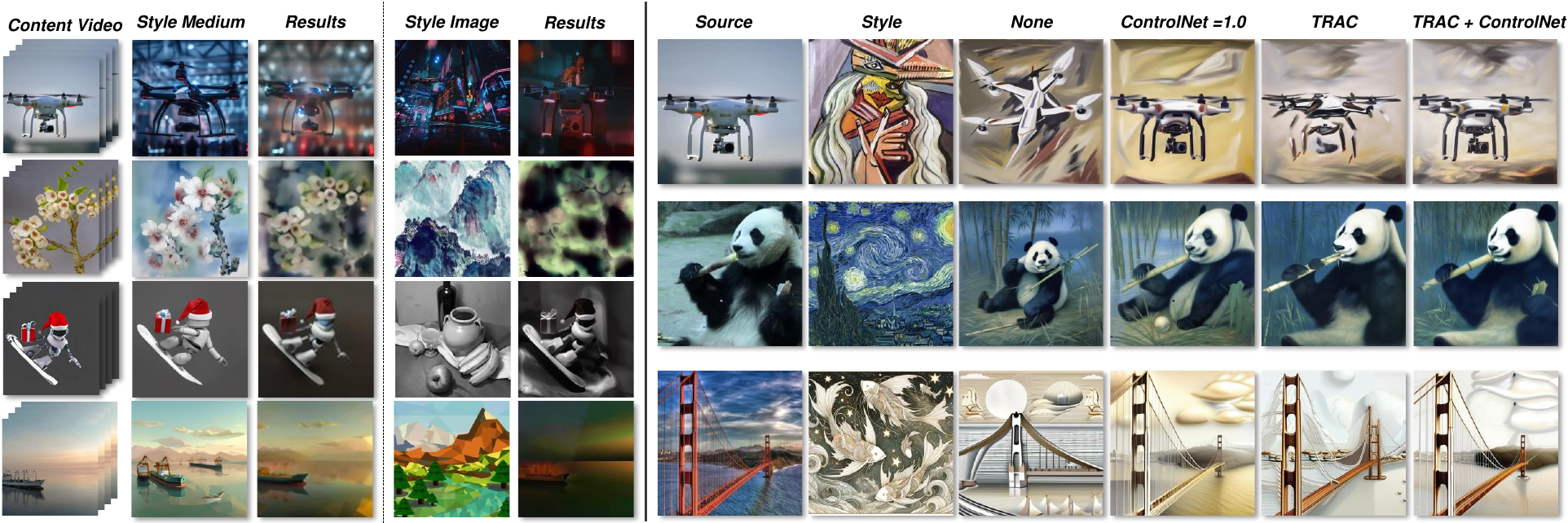} 
\caption{The left side compares the results generated using the style medium with those obtained directly from the style image. The right side provides results with different combinations of content control methods applied.}
\label{fig: restrict}
\end{figure*}

\textbf{Style Meidum}
The concept of a style medium is essential in preventing content leakage, but could stricter regulation of shape and content eliminate its necessity? To assess this, we replaced the style medium with the original style reference image and applied high control strength to regulate content leakage.
As shown on the left of \cref{fig: restrict}, replacing the style medium with high structure and content control introduces significant artifacts in stylized videos, such as color distortion and blurriness. Despite this, content leakage still persists. We attribute these artifacts to the lack of alignment between style and content, as the style medium serves as a crucial intermediary.
Without this bridge, the connection between the style image and content video remains weak, making it difficult for the cross-attention mechanism to establish high-level semantic connections. As a result, the mechanism primarily focuses on low-level features like color and texture, leading to suboptimal style authenticity.

Comparing the previous results with those generated without the style medium confirms its crucial role in facilitating style transfer between otherwise unconnected candidates. Acting as a bridge, it enables seamless integration of style information from the reference image into the content video while preserving content similarity. Furthermore, the style medium mitigates challenges related to content preservation. In its absence, excessive content leakage can dominate the entire image, disrupting the balance between content and style. By regulating the flow of style information, the style medium ensures that content integrity is maintained while still achieving effective style transfer.




\textbf{TRAC and ControlNet}
TRAC effectively controls both content and layout, with ControlNet \cite{zhang2023addingconditionalcontroltexttoimage} enhancing fine-grained structural control. Ablation studies were conducted to assess their individual contributions.
As shown on the right side of \cref{fig: restrict}, ControlNet effectively controls structure, especially in maintaining precise shapes. The control effect depends on the provided control information, such as Canny edges. However, content leakage remains evident in areas not explicitly defined by the edges. In \cref{fig: restrict}, although the edges remain unchanged, the panda is noticeably deformed compared to the source, leading to significant semantic alteration.
The control primarily emphasizes structural details rather than content and layout, and its rigid nature often limits the adaptation of edges to the target art style.
In contrast, TRAC effectively manages global content and layout, scaling it up results in the minimization of content leakage. TRAC provide a less rigid type of control, it allows structural elements to be slightly altered, providing more flexibility for authentic style changes.

Given that style transfer is an inherently subjective task, ControlNet is incorporated into the final framework. The combination of TRAC and ControlNet can provide great flexibility.
By combining the global content and layout control from TRAC with the low-level structure preservation capabilities of ControlNet, our framework supports a broad spectrum of stylized outputs, ranging from precise content restoration to the generation of visually striking results with vibrant and expressive styles.




\begin{table}
\centering
\small
\begin{tabular}{c c c c c c}
\hline
Metric & long.short &  long.long  & short.long & short.short \\ 
\hline
\hline
 $\downarrow$ ArtFID & 36.63 &  37.73 & 37.64 & 37.57  \\ 
 $\downarrow$ FID & 24.41 & 24.30 & 24.12 & 24.20\\ 
 $\downarrow$ LPIPS & 0.4413  & 0.4917  & 0.4985 & 0.4908   \\ 
 \hline 
 \\
\end{tabular}
\caption{Ablation on different prompt choices.}
\label{tab: ablation prompt}
\end{table}

\textbf{Prompt choice}
An ablation study is conducted to understand the difference in each prompt choice. We adopt different combinations in using the long prompt and the short prompt. As shown in \cref{tab: ablation prompt}, the results in other prompt combinations all yield suboptimal results in LPIPS, which signify the inconsistency in the content. 
We conjecture that the generation of the style medium needs to utilize the long prompt to encapsulate all possible elements into the style medium. However, in the generation of the later stage, the structural details contained in the long prompt may conflict with the content control we enforced, resulting in suboptimal content consistency. The changes in the FID score are minimal, which we attribute to normal fluctuations.

\section{Conclusion}
\label{sec:con}

In conclusion, our research tackles the key challenges of content leakage and style misalignment in video style transfer. By introducing the style medium, we enable a more refined transfer of stylistic features, reducing content leakage while achieving visually compelling results. Our innovative inversion-free attention control method, TRAC, further enhances this process by providing explicit content layout regulation without relying on inversion techniques.
The proposed framework not only surpasses previous methods in maintaining coherence and authenticity but also offers greater flexibility.
TRAC, combined with ControlNet, enables multi-layer control for more precise and flexible adjustments.
Regarding potential impact, the style medium concept can be seamlessly integrated into existing methods to enhance their performance. Furthermore, TRAC, a more efficient alternative to DDIM inversion, can be directly incorporated into current multi-path style transfer frameworks and broader image editing techniques, offering a more efficient solution.

{
    \small
    \bibliographystyle{ieeenat_fullname}
    \bibliography{main}
}
\end{document}